%% file: ms.tex
\documentclass[10pt,twocolumn,letterpaper]{article}

\usepackage{cvpr}
\usepackage{times}
\usepackage{epsfig}
\usepackage{graphicx}
\usepackage{amsmath}
\usepackage{amssymb}

\usepackage{gensymb}
\usepackage{makecell}
\usepackage{hhline}
\usepackage{subcaption}
\usepackage{microtype}

\usepackage[skip=2pt]{caption}
\usepackage[toc,page]{appendix}

\newcommand{\KS}[1]{\textcolor{blue}{\textbf{KS: #1}}}

\newcommand{\mc}[1]{\mathcal{#1}}

\newcommand{\la}[0]{\leftarrow}

\newcommand{\spaceunderfigures}{-0.15in}

\newcommand{\boldstart}[1]{\noindent\textbf{#1}}
\newcommand{\boldstartspace}[1]{\vspace{0.1in}\noindent\textbf{#1}}
\newcommand{\comment}[1]{}

\usepackage[pagebackref=true,breaklinks=true,letterpaper=true,colorlinks,bookmarks=false]{hyperref}

\cvprfinalcopy 

\def\cvprPaperID{7020}

\ifcvprfinal\fi
\begin{document}

\title{Deep 3D Capture: Geometry and Reflectance from Sparse Multi-View Images}


\author{
\hspace{0.5cm}
Sai Bi\textsuperscript{1} \hspace{0.5cm}
Zexiang Xu\textsuperscript{1} \hspace{0.5cm}
Kalyan Sunkavalli\textsuperscript{2} \hspace{0.5cm}
David Kriegman\textsuperscript{1} \hspace{0.5cm}
Ravi Ramamoorthi\textsuperscript{1} \hspace{0.4cm} 
\\
\textsuperscript{1}UC San Diego\hspace{1cm}
\textsuperscript{2}Adobe Research 
}

\maketitle
\thispagestyle{empty}

\begin{abstract}
We introduce a novel learning-based method to reconstruct the high-quality geometry and complex, spatially-varying BRDF of an arbitrary object from a sparse set of only six images captured by wide-baseline cameras under collocated point lighting. 
We first estimate per-view depth maps using a deep multi-view stereo network; these depth maps are used to coarsely align the different views.
We propose a novel multi-view reflectance estimation network architecture that is trained to pool features from these coarsely aligned images and predict per-view spatially-varying diffuse albedo, surface normals, specular roughness and specular albedo.
Finally, we fuse and refine these per-view estimates to construct high-quality geometry and per-vertex BRDFs.
We do this by jointly optimizing the latent space of our multi-view reflectance network to minimize the photometric error between images rendered with our predictions and the input images.
While previous state-of-the-art methods fail on such sparse acquisition setups, we demonstrate, via extensive experiments on synthetic and real data, that our method produces high-quality reconstructions that can be used to render photorealistic images. 

\end{abstract}
\input{Introduction}

\input{RelatedWorks}

\input{Algorithm}

\input{Results}

\input{Conclusions}


\newpage
{\small
\bibliographystyle{ieee_fullname}
\bibliography{egbib}
}

\newpage
\begin{appendices}
\input{network.tex}

\end{appendices}

\end{document}

%% file: Introduction.tex
\vspace{-0.3cm}
\section{Introduction}\label{sec:intro}
\begin{figure}[t]
    \centering
    \includegraphics[width=\linewidth]{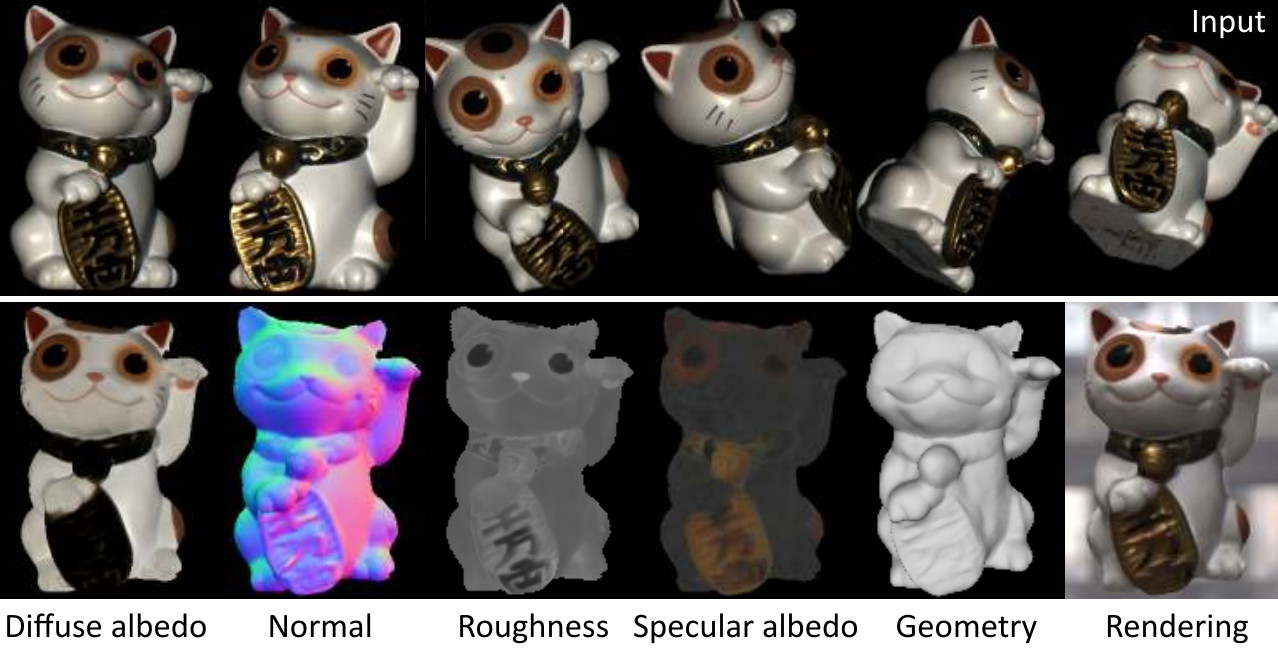}
    \caption{From six wide-baseline input images of an object captured under collocated point lighting (top row), our method reconstructs high-quality geometry and spatially-varying, non-Lambertian reflectance (bottom row,
    a tone mapping is performed on specular albedo to make it more visible), allowing us to re-render the captured object under novel viewpoint and illumination (bottom, right).
    }
    \vspace{-0.5cm}
    \label{fig:teaser}
\end{figure}

Reconstructing the 3D geometry and reflectance properties of an object from 2D images has been a long-standing problem in 
computer vision and graphics, with applications including 3D visualization, relighting, and augmented and virtual reality. 
Traditionally this has been accomplished using complex acquisition 
systems~\cite{baek2018simultaneous, holroyd2010coaxial,tun2013acquiring,wu2015simultaneous,zhou2016sparse} 
or multi-view stereo (MVS) methods~\cite{furukawa2015multi, schoenberger2016mvs} 
applied to dense image sets~\cite{nam2018practical, xia2016recovering}. The acquisition requirements for these methods significantly limits their practicality.  
Recently, deep neural networks have been proposed for material estimation from 
a single or a few images. 
However, many of these methods are restricted to estimating the spatially-varying BRDF (SVBRDF) of planar samples~\cite{deschaintre2018single, gao2019deep, li2018materials}.
Li et al.~\cite{li2018learning} demonstrate shape and reflectance reconstruction from a single image, but their reconstruction quality is limited by their single image input.

\begin{figure*}[t]
    \centering
    \includegraphics[width=\textwidth]{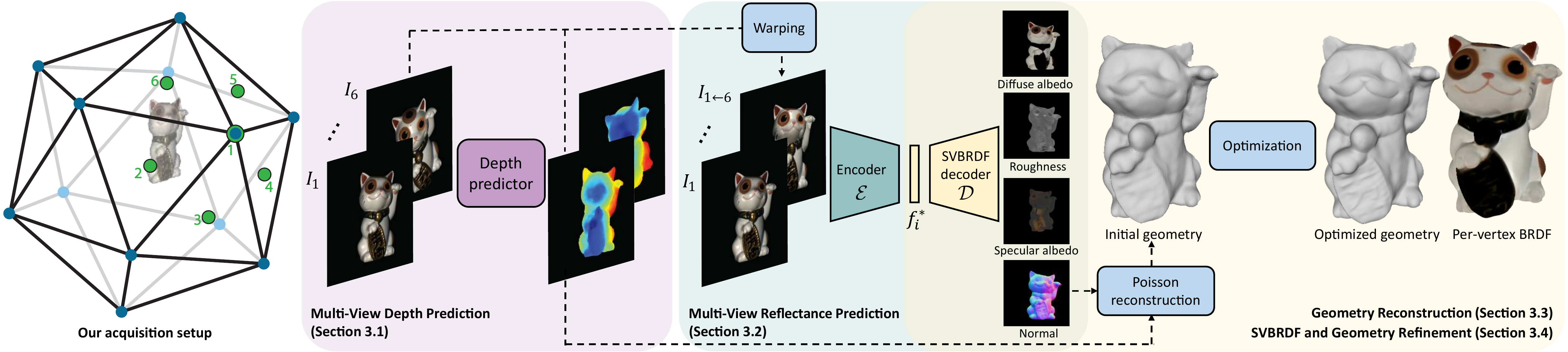}
    \caption{Our acquisition setup (leftmost figure) and framework. 
        We capture six images with collocated cameras and lights placed at a vertex (\emph{green} circle $1$)
        and five adjoining face centers (\emph{green} circle $2$-$6$) of an icosahedron. 
        Using the six images, we predict per-view depth (\emph{red block}). 
        We warp the input images using the predicted depths and pass them to a multi-view SVBRDF estimation network to get per-view SVBRDFs (\emph{blue block}). Finally, we reconstruct 3D geometry from the estimated 
        depth and normals, and perform a joint optimization to get refined geometry and per-vertex BRDFs (\emph{yellow block}).
    }
    \label{fig:pipeline}
    \vspace{\spaceunderfigures}
\end{figure*}

Our goal is to enable practical \emph{and} high-quality shape and appearance acquisition. 
To this end, we propose using a simple capture setup: a sparse set of six cameras---placed at one vertex and the centers of the adjoining faces of a regular icosahedron, forming a $60^{\circ}$ cone---with collocated point lighting (Fig.~\ref{fig:pipeline} left). 
Capturing six images should allow for better reconstruction compared to single image methods. 
However, at such wide baselines, the captured images have few correspondences and severe occlusions, making it challenging to fuse information across viewpoints.

As illustrated in Fig.~\ref{fig:pipeline}, we propose a two-stage approach to address this problem. First, we design \textbf{multi-view geometry and reflectance estimation networks that regress the 2D depth, normals and reflectance for each input view by robustly aggregating information across all sparse viewpoints.}
We estimate the depth for each input view using a deep multi-view stereo network~\cite{xu2019deep, yao2018mvsnet} (Sec.~\ref{sec:depth}). 
Because of our sparse capture, these depth maps contain errors and cannot be used to accurately align the images to estimate per-vertex BRDFs~\cite{nam2018practical, zhou2016sparse}.
Instead, we use these depth maps to warp the images to one viewpoint and use a \emph{novel deep multi-view reflectance estimation network} to estimate per-pixel normals and reflectance (parameterized by diffuse albedo, specular albedo and roughness in a simplified Disney BRDF model~\cite{karis2013real}) 
for that viewpoint (Sec.~\ref{sec:svbrdf}).
This network extracts features from the warped images, aggregates them across viewpoints using max-pooling, and decodes the pooled features to estimate the normals and SVBRDF for that viewpoint.
This approach to aggregate multi-view information leads to more robust reconstruction than baseline approaches like a U-Net architecture~\cite{ronneberger2015u}, 
and we use it to recover normals and reflectance for each view. 

Second, we propose a \textbf{novel method to fuse these per-view estimates into a single mesh with per-vertex BRDFs using optimization in a learnt reflectance space}.
First, we use Poisson reconstruction~\cite{kazhdan2006poisson} to construct a mesh from the estimated per-view depth and normal maps (Sec.~\ref{sec:geometry}).
Each mesh vertex has multiple reflectance parameters corresponding to each per-view reflectance map, and we fuse these estimates to reconstruct object geometry and reflectance that will \emph{accurately reproduce the input images}.
Instead of optimizing the per-vertex reflectance parameters, which leads to outliers and spatial discontinuities, we optimize the \emph{the latent features of our multi-view reflectance estimation network} (Sec.~\ref{sec:opt}).
We pass these latent features to the reflectance decoder to construct per-view SVBRDFs, fuse them using per-vertex blending weights, and render them to compute the photometric error for all views.
This entire pipeline is differentiable, allowing us to backpropagate this error and iteratively update the reflectance latent features and per-vertex weights till convergence.
This process refines the reconstruction to best match the specific captured images, while leveraging the priors learnt by our reflectance estimation network.

We train our networks with a large-scale synthetic dataset comprised of procedurally generated shapes 
with complex SVBRDFs~\cite{xu2019deep,xu2018deep} and rendered using a physically-based renderer. 
While our method is trained with purely synthetic data, it generalizes well to real scenes. 
This is illustrated in Figs.~\ref{fig:teaser} and \ref{fig:real-results}, where we are able to reconstruct real objects with complex geometry and non-Lambertian reflectance.
Previous state-of-the-art methods, when applied to sparse input images for such objects, produce incomplete, noisy geometry and erroneous reflectance estimates (Figs.~\ref{fig:comp-colmap} and \ref{fig:comp-nam}).
In contrast, our work is the first to reconstruct detailed geometry and high-quality reflectance from sparse multi-view inputs, allowing us to render photorealistic images under novel view and lighting.

%% file: RelatedWorks.tex
\section{Related Works}\label{sec:related}

\boldstart{3D reconstruction.}
To reconstruct 3D geometry from image sets, traditional methods~\cite{furukawa2009accurate, langguth2016smvs, schoenberger2016mvs}
find correspondences between two or more images utilizing specific image features. Such methods are
sensitive to illumination changes, non-Lambertian reflectance and textureless surfaces. The existence 
of multiple points with similar matching costs also require these methods to have a large number of 
images to get high-quality reconstructions (we refer the interested readers to~\cite{furukawa2009accurate} for more details). 
In contrast, our method reconstructs high-quality geometry
for complex real scenes from an order of magnitude fewer images. 

Recently, numerous learning-based methods have been proposed to reconstruct 3D shape using various geometric representations, 
including regular volumes \cite{ji2017surfacenet,richter2018matryoshka,wu2017marrnet}, 
point clouds \cite{achlioptas2018learning,wang2018mvpnet} and depth maps~\cite{huang2018deepmvs,yao2018mvsnet}.
These methods cannot produce high-resolution 3D meshes.
We extend recent learning-based MVS frameworks \cite{xu2019deep, yao2018mvsnet} 
to estimate depth from sparse multi-view images of objects with complex reflectance.
We combine this depth with estimated surface normals to reconstruct 3D meshes with fine details.

\boldstartspace{SVBRDF acquisition.} 
SVBRDF acquisition is a challenging task that often requires a dense input image set \cite{dong2014appearance,nam2018practical,xia2016recovering}.
Many methods utilize sophisticated hardware \cite{matusik2003data} or light patterns \cite{holroyd2010coaxial,kang2018efficient,tun2013acquiring}.
Reconstruction from sparse images has been demonstrated for planar objects \cite{aittala2015two,li2018materials,xu2016minimal}, 
and known geometry \cite{zhou2016sparse}. 
In contrast, we reconstruct the geometry and complex reflectance of arbitrary objects from a sparse set of six input images.

Photometric stereo methods have been proposed to reconstruct arbitrary shape and SVBRDFs~\cite{alldrin2008photometric,goldman2009shape}; 
however, they focus on single-view reconstruction and require hundreds of images.
Recent works~\cite{hui2017reflectance,nam2018practical} utilize images captured by a collocated camera-light setup for shape and SVBRDF estimation. 
In particular, Nam et al.~\cite{nam2018practical} capture more than sixty images and use multi-view reconstruction and physics-based optimization to recover geometry and reflectance.
In contrast, by designing novel deep networks, we are able to reconstruct objects from only six images.

Learning-based methods have been applied for normal and SVBRDF acquisition.
Deep photometric stereo methods reconstruct surface normals from tens to hundreds of images~\cite{chen2019self,chen2018psfcn} but they do not address reflectance or 3D geometry estimation.
Most deep SVBRDF acquisition methods are designed for planar samples ~\cite{aittala2016neural,deschaintre2018single,deschaintre2019multi,gao2019deep,li2017appearance,li2018materials}.
Some recent multi-image SVBRDF estimation approaches pool latent features from multiple views~\cite{deschaintre2019multi} and use latent feature optimization~\cite{gao2019deep} but they only handle planar objects.
Li et al.~\cite{li2018learning} predict depth and SVBRDF from a single image; 
however, a single input does not provide enough information to accurately reconstruct geometry and reflectance. 
By capturing just six images, our approach generates significantly higher quality results.

%% file: Algorithm.tex
\section{Algorithm}\label{sec:algorithm}

Our goal is to accurately reconstruct the geometry and SVBRDF of an object 
with a simple acquisition setup. 
Recent work has utilized collocated point illumination for reflectance estimation from 
a sparse set of images~\cite{aittala2016neural,aittala2015two,deschaintre2018single,li2018materials}; 
such lighting minimizes shadows and induces high-frequency effects like specularities, making reflectance estimation easier.
Similarly, Xu et al.~\cite{xu2019deep} demonstrate novel view synthesis from sparse multi-view images of a scene captured under a single point light.

Motivated by this, we utilize a similar capture system as Xu et al.---six cameras placed at one vertex of a regular icosahedron and the centers of the five faces adjoining that vertex. 
Unlike their use of a single point light for all images, we capture each image under a point light (nearly) collocated with the corresponding camera (see Fig.~\ref{fig:pipeline} left). 
The setup is calibrated giving us a set of $n=6$ input images, $\{I_i\}_{i=1}^n$ with the corresponding camera calibration. 
This wide baseline setup---with an angle of $37^{\circ}$ between the center and boundary views---makes it possible to image the entire object with a small set of cameras.
In the following, we describe how we reconstruct an object from these sparse input images.

\subsection{Multi-View Depth Prediction}\label{sec:depth}
Traditional MVS methods depend on hand-crafted features such as Harris descriptors to find correspondence between views. 
Such features are not robust to illumination changes or non-Lambertian surfaces, making them unusable for our purposes.
In addition, due to the sparse inputs and large baselines, parts of the object may be visible in as few as two views. 
These factors cause traditional MVS methods to fail at finding accurate correspondences, and thus fail to reconstruct high-quality geometry.

Instead, we make use of a learning-based method to estimate the depth. 
Given the input images $\{I_i\}_{i=1}^n$, we estimate the depth map $D_i$ for view $i$. Similar to recent works on learning-based MVS~\cite{im2019dpsnet, xu2019deep, yao2018mvsnet}, our network consists of two components: a feature extractor $\mc{F}$ and a correspondence predictor $\mc{C}$. 
The feature extractor is a 2D U-Net~\cite{ronneberger2015u} that extracts a $16$-channel feature map for each image $I_i$. 
To estimate the depth map at $I_i$, we warp the feature maps of all views to view $i$ using a 
set of $128$ pre-defined depth levels, and build a 3D plane sweep volume~\cite{collins1996space} by 
calculating the variance of feature maps over views. 
The 3D volume is further fed to the correspondence predictor $\mc{C}$ 
that is a 3D U-Net
to predict the probability of each depth level. 
We calculate the depth as a probability-weighted sum of all depth levels. 
The training loss is defined as the $L_1$ loss between predicted depths and ground truth depths. 
By learning the feature representations and correspondence, the proposed framework is more robust to illumination changes and specularities, thus producing more accurate pixel-wise depth predictions than traditional methods. 

While such networks are able to produce reasonable depth, the recovered depth has errors in textureless regions. 
To further improve the accuracy, we add a guided filter module~\cite{wu2018fast} to the network, which includes a guided map extractor $\mc{G}$ as well as a guided layer $g$.
Let the initial depth prediction at view $i$ be $D_i'$. 
The guided map extractor $\mc{G}$ takes image $I_i$ as input and learns a guidance map $\mc{G}(I_i)$. 
The final depth map is estimated as:
\begin{align}
    D_i = g(\mc{G}(I_i), D_i').
\end{align}
The training loss is defined as the $L_1$ distance between predicted depths and ground truth depths. All components are trained jointly in an end-to-end manner. 


\subsection{Multi-View Reflectance Prediction}\label{sec:svbrdf}

Estimating surface reflectance from sparse images is a highly under-constrained problem. 
Previous methods either assume geometry is known~\cite{aittala2016neural,aittala2015two,li2018materials,deschaintre2018single} or can be reconstructed with specific devices~\cite{holroyd2010coaxial} or MVS~\cite{nam2018practical}. In our case, accurate geometry cannot be reconstructed from sparse inputs with traditional MVS methods.
While our learning-based MVS method produces reasonable depth maps, they too have errors,
making it challenging to use them to align the images and estimate per-vertex SVBRDF.
Instead, for each input image $I_i$, we first estimate its corresponding normals, $N_i$, and SVBRDF, represented by diffuse albedo $A_i$, specular roughness $R_i$ and specular albedo $S_i$.

To estimate the SVBRDF at view $i$, we warp all input images to this view using 
predicted depths $D_i$.
One approach for multi-view SVBRDF estimation could be to feed this stack of warped images to a convolutional neural network like the commonly used U-Net~\cite{li2018materials, ronneberger2015u}. 
However, the inaccuracies in the depth maps lead to misalignments in the warped images, especially in occluded regions, and this architecture is not robust to these issues.

\begin{figure}[t]
    \centering
    \includegraphics[width=\linewidth]{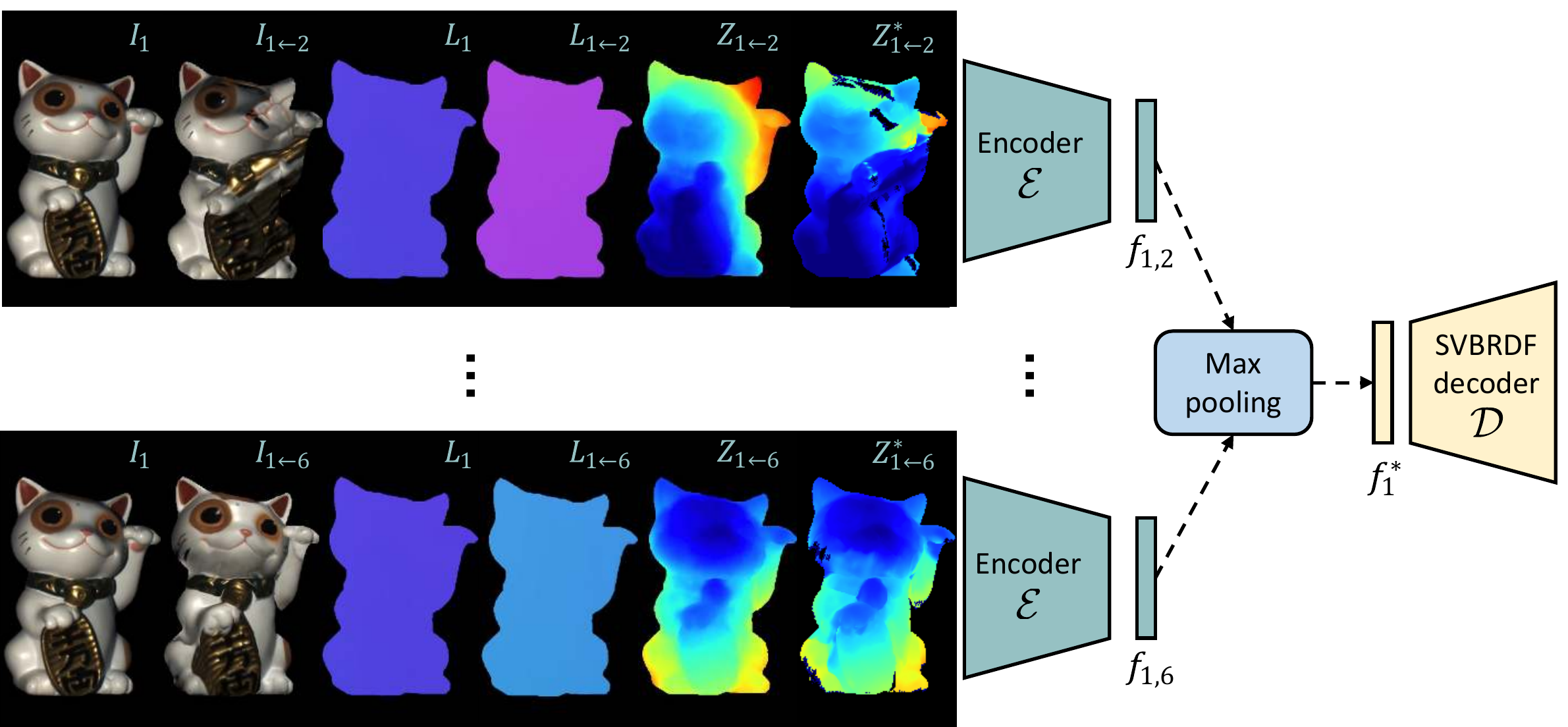}
    \caption{Our multi-view SVBRDF estimation network. An encoder extracts features from reference and warped image pairs. These features are max-pooled to 
        get a single reference-view feature map, which is decoded to predict that view's SVBRDF. Note the errors in the warped images; max-pooling mitigates their effect on the output SVBRDF.
    }
    \label{fig:svbrdf-network}
    \vspace{\spaceunderfigures}
\end{figure}

We propose a novel architecture that is robust to depth inaccuracies and occlusions. 
As shown in Fig.~\ref{fig:svbrdf-network}, our network comprises a Siamese encoder~\cite{chopra2005learning}, $\mc{E}$, and a decoder, $\mc{D}$, with four branches for the four SVBRDF components.  
To estimate the SVBRDF at a reference view $i$, the encoder processes $n$ pairs of inputs, each pair including image $I_i$ as well as the warped image
$I_{i \la j}$, where we warp image $I_j$ at view $j$ to the reference view $i$ using 
the predicted depth $D_i$.
To handle potential occlusions, directly locating occluded regions in the warped images using predicted depths and masking them out is often not feasible due to inaccurate depths. 
Instead we keep the occluded regions in the warped images and include the depth information
in the inputs, allowing the network to learn which parts are occluded. 

To include the depth information, 
we draw inspiration from the commonly used shadow mapping technique~\cite{williams1978casting}. The depth input consists of two components: for each pixel in view $i$, we calculate its depths $Z_{i\la j}$ in view $j$;
we also sample its depth $Z^*_{i \la j}$ from the depth map $D_j$ by finding its projections 
on view $j$. 
Intuitively if $Z_{i \la j}$ is larger than $Z^*_{i \la j}$, then the pixel is 
occluded in view $j$; otherwise it is not occluded. 
In addition, for each pixel in the reference view $i$, we also include the lighting directions $L_i$ of the light at view $i$, as well as the lighting direction of the light at view $j$, denoted as $L_{i\la j}$.
We assume a \textit{point light} model here. Since the light is collocated with the camera, by including the lighting direction we are also including the viewing direction of each pixel in the inputs.
All directions are in the coordinate system of the reference view. Such cues are critical for networks to infer surface normals using photometric information. 
Therefore, the input for a pair of views $i$ and $j$ is:
\begin{align}
    H_{i, j} = \{I_i, I_{i\la j}, Z_{i \la j}, Z^*_{i \la j}, L_i, L_{i \la j}\}.
    \label{eqn:encoder-input}
\end{align}
The input contains $14$ channels in total, and there are a total of $n$ such inputs. We feed all the inputs to the encoder network $\mc{E}$ and get the intermediate 
features $f_{i, j}$. 
All these intermediate features are aggregated with a max-pooling layer yielding a common feature representation for view $i$, $f_i^*$:
\begin{align}
    f_{i, j} &= \mc{E}(H_{i,j}) \\
    f_i^* &= \text{max-pool}(\{f_{i,j}\}_{j=1}^{n}) 
    \label{eqn:latent}
\end{align}
$f_i^*$ is fed to the decoder to predict each SVBRDF component for view $i$:
\begin{align}
    A_i, N_i, R_i, S_i = \mc{D}(f_i^*)
    \label{eqn:decoder}
\end{align}
Compared to directly stacking all warped images together, our proposed network architecture works on pairs of input 
images and aggregates features across views using a max-pooling layer. The use of max-pooling 
makes the network more robust to occlusions and misalignments caused by depth inaccuracies and produces more accurate results (see Tab.~\ref{table:comp-synthetic}). It also makes the network invariant to the number and order of the input views, a fact that could be utilized for unstructured capture setups.
The training loss $\mc{L}$ of the network is defined as:
\begin{align} 
    \mc{L} = \mc{L}_A + \mc{L}_N + \mc{L}_R + \mc{L}_S + \mc{L}_{I}
\end{align}
where the first four terms are the $L_2$ losses for each SVBRDF component, and $\mc{L}_I$ is the $L_2$ loss 
between input images and rendered images generated with our predictions. 

\subsection{Geometry Reconstruction}\label{sec:geometry}

\comment{
Previous methods reconstruct 3D geometries by directly predicting
occupancy grids~\cite{kar2017learning}, signed distance volumes~\cite{riegler2017octnetfusion}, implicit 
signed distance functions~\cite{park2019deepsdf, saito2019pifu} or 
triangle meshes~\cite{wang2018pixel2mesh} from multi-view images or depth maps.
Such methods suffer from limited resolutions and cannot recover fine-grained details of the 
objects. Moreover, they are restricted to certain categories of objects and do not generalize to objects of unseen categories. \KS{not sure this paragraph is necessary}
}

The previous multi-view depth and SVBRDF estimation networks give us per-view depth and normal maps at full-pixel resolution.
We fuse these per-view estimates to reconstruct a single 3D geometry for the object. 
We first build a point cloud from the depth maps, by generating 3D points from each pixel in every per-view depth map.
For each point, we also get its corresponding normal from the estimated normal maps. 
Given this set of 3D points with surface normals, we perform a Poisson reconstruction~\cite{kazhdan2013screened} to reconstruct the fused 3D geometry. 
The initial point clouds may contain outliers due to inaccuracies in the depth maps.
To get rid of undesired structures in the output geometry, we generate a coarse initial geometry by setting the depth of the spatial octree in Poisson reconstruction to $7$---corresponding to 
an effective voxel resolution of $128^3$. 
We refine this initial geometry in the subsequent stage.
Compared to learning-based 3D reconstruction methods that directly generate geometry (voxel grids~\cite{kar2017learning,riegler2017octnetfusion}, implicit 
functions~\cite{park2019deepsdf, saito2019pifu} or 
triangle meshes~\cite{wang2018pixel2mesh}) from images, this approach generalizes to arbitrary shapes and produces more detailed reconstructions. 

\subsection{SVBRDF and Geometry Refinement}\label{sec:opt}

Given the initial coarse geometry as well as the per-view SVBRDF predictions, we aim to construct a detailed 3D mesh with per-vertex BRDFs.
For each vertex, a trivial way to get its BRDF is to blend the predicted SVBRDFs across views using pre-defined weights such as the dot product of the viewing directions and surface normals. 
However, this leads to blurry results (Fig.~\ref{fig:comp-direct-opt}), due to the inconsistencies in the estimated SVBRDFs and the geometry.
Also note that our SVBRDF predictions are computed from a single feed-forward network pass, and are not guaranteed to reproduce the captured input images exactly because the network has been trained to minimize the reconstruction loss on the entire training set and not this specific input sample.

We address these two issues with a novel rendering-based optimization that estimates per-vertex BRDFs that minimize the error between rendering the predicted parameters and the captured images.
Because of the sparse observations, independently optimizing per-vertex BRDFs leads to artifacts such as outliers and spatial discontinuities, as shown in Fig.~\ref{fig:comp-direct-opt}.
Classic inverse rendering methods address this using hand-crafted priors.
Instead, we optimize the per-view feature maps $f_i^*$ that are initially predicted from our SVBRDF encoder ( Eqn.~\ref{eqn:latent}).
These latent features, by virtue of the training process, capture the manifold of object reflectances, and generate spatially coherent per-view SVBRDFs when passed through the decoder, $\mc{D}$ (Eqn.~\ref{eqn:decoder}).
\emph{Optimizing in this feature space allows us to adapt the reconstruction to the input images, while leveraging the priors learnt by our multi-view SVBRDF estimation network.}

\boldstartspace{Per-vertex BRDF and color.}
For each vertex $v_k$, we represent its BRDF $b_k$ as a weighted average of the BRDF predictions from multiple views:
\begin{align}
b_k = \sum_{i=1}^{n} w_{k, i} \mc{D}(p_{k,i};f_i^*),
\end{align}
where $p_{k,i}$ is the corresponding pixel position of $v_k$ at view $i$, $\mc{D}(p_{k,i};f_i^*)$ represents the SVBRDF prediction at $p_{k,i}$ from view $i$ by processing $f_i^*$ via the decoder network $\mc{D}$, and $w_{k, i}$ are the per-vertex view blending weights. 
The rendered color of $v_k$ at view $i$ is calculated as: 
\begin{align}
I^*_i(p_{k,i}) = \Theta(b_k, L_i(p_{k,i})),
\end{align}
where $L_i(p_{k,i})$ is the lighting direction and also the viewing direction of vertex $v_k$ at view $i$, and $\Theta$ is the rendering equation. 
We assume a point light source collocated with the camera (which allows us to ignore shadows), and only consider direct illumination in the rendering equation. 

\boldstartspace{Per-view warping.}
Vertex $v_k$ can be projected onto view $i$ using the camera calibration; we refer to this projection as $u_{k,i}$. However, the pixel projections onto multiple views might be inconsistent due to inaccuracies in the reconstructed geometry. Inspired by Zhou et al.~\cite{zhou2014color}, we apply a non-rigid warping to each view to better align the projections. 
In particular, for each input view, we use a $T\times T$ grid with $C=T^2$ control points ($T=11$ in our experiments) to construct a smooth warping field over the image plane.
Let $t_{i,c}$ be the translation vectors of control points at view $i$. 
The resulting pixel projection, $p_{k,i}$, is given by:
\begin{align}
p_{k,i} = u_{k,i} + \sum_{c=1}^{C} \theta_c (u_{k,i}) t_{i, c},
\end{align}
where $\theta_c$ returns the bilinear weight for a control point $t_{i, c}$ at pixel location $u_{k,i}$. 

\boldstartspace{SVBRDF optimization.}
We optimize per-view latent features $f_i^*$, per-vertex blending weights $w_{k, i}$ and per-view warping fields $t_{i, c}$ to reconstruct the final SVBRDFs.
The photometric consistency loss between the rendered colors and ground truth colors for all $K$ vertices is given by:
\begin{align}
\label{equ:opt}
    E_{\text{photo}} (f^*_i, w, t) 
    = \frac{1}{n\cdot K}\sum_{k=1}^{K}\sum_{i=1}^{n} ||I^*_i(p_{k,i}) - I_i(p_{k,i})||_2^2. \nonumber 
\end{align}
We clamp the rendered colors to the range of $[0,1]$ before calculating the loss.
To prevent the non-rigid warping from drifting, we also add an $L_2$ regularizer to penalize the norm of the translation vectors:
\begin{align}
    E_{\text{warp}}(t)= \frac{1}{n\cdot C}\sum_{i=1}^{n}\sum_{c=1}^{C} ||t_{i,c}||_2^2.
\end{align}
Therefore the final energy function for the optimization is:
\begin{align}
    E = E_\text{photo}(f^*, w, t) + \lambda E_{\text{warp}}(t).
\end{align}
We set $\lambda$ to $100$ , and optimize the energy function with 
Adam optimizer~\cite{kingma2014adam} with a learning rate of $0.001$.


\boldstartspace{Geometry optimization.}
We use the optimized per-vertex normal, $n_k$, to update the geometry of the object by 
re-solving the Poisson equation (Sec.~\ref{sec:geometry}). 
Unlike the initial geometry reconstruction, we set the depth of the spatial octree to $9$---corresponding to a voxel resolution of $512^3$---to better capture fine-grained details of the object.
We use this updated geometry in subsequent SVBRDF optimization iterations. 
We update the geometry once for every $50$ iterations of SVBRDF optimization, and we perform $400-1000$ iterations for the SVBRDF optimization.

\boldstartspace{Per-vertex refinement.} 
The bottleneck in our multi-view SVBRDF network---that we use as our reflectance representation---may cause a loss of high-frequency details in the predicted SVBRDFs. 
We retrieve these details back by directly optimizing the BRDF parameters $b_k$ of each vertex to minimizing the photometric loss in Eqn.~\eqref{equ:opt}. 
Note that after the previous optimization, the estimated BRDFs have already converged to good results and the rendered images are very close to the input images.
Therefore, in this stage, we use a small learning rate ($0.0005$), and perform the optimization for a small number ($40-100$) of iterations.

%% file: Results.tex
\section{Implementation and Results}\label{sec:results}

\boldstart{Training data.}
We follow Xu et al.~\cite{xu2019deep} and procedurally generate complex scenes
by combining $1$ to $5$ primitive shapes such as cylinders and cubes displaced by random height maps.
We generate $20,000$ training and $400$ testing scenes.
We divide the high-quality materials from the Adobe Stock dataset\footnote{\url{https://stock.adobe.com/search/3d-assets}} into a training and testing set,
and use them to texture the generated scenes separately. For each scene, 
following the setup discussed in Sec.~\ref{sec:intro},
we render the 6 input view images with a resolution of $512 \times 512$ using a custom Optix-based global illumination 
renderer with 1000 samples per pixel. We also render the ground truth depth, normals, and SVBRDF components for each view. 

\boldstart{Network architecture.} For depth estimation, we use a 2D U-Net architecture~\cite{ronneberger2015u} 
for the feature extractor, $\mc{F}$, and guidance map extractor, $\mc{G}$. Both networks have $2$ downsampling/upsampling blocks. The correspondence predictor $\mc{C}$ is a 3D U-Net 
with $4$ downsampling/upsampling blocks.  
For multi-view SVBRDF estimation, both the encoder $\mc{E}$ and decoder $\mc{D}$ are 2D CNNs, with $3$ downsampling layers in $\mc{E}$ and $3$ upsampling layers in $\mc{D}$.
Note that we \emph{do not} use skip connections in the SVBRDF network; this forces the latent feature to learn a meaningful reflectance space and allows us to optimize it in our refinement step.
We use group normalization~\cite{wu2018group} in all networks.
We use a differentiable rendering layer that computes local shading under point lighting without considering visibility or global illumination. This is a reasonable approximation in our collocated lighting setup.
For more details, please refer to the supplementary document.

\boldstart{Training details.}  All the networks are trained with 
the Adam optimizer~\cite{kingma2014adam} for $50$ epochs with a learning rate of $0.0002$. 
The depth estimation networks are trained on cropped patches of $64 \times 64$ 
with a batch size of $12$, and the SVBRDF estimation networks are trained 
on cropped $320 \times 320$ patches with a batch size of $8$. Training took around four days
on $4$ NVIDIA Titan 2080Ti GPUs.


\boldstart{Run-time.} 
Our implementaion has not been optimized for the best timing efficiency.  In practice, our method takes around $15$
minutes for full reconstruction from images with a resolution of $512 \times 512$, where 
most of the time is for geometry fusion and optimization. 


\subsection{Evaluation on Synthetic Data}

We evaluate our max-pooling-based multi-view SVBRDF estimation network on our synthetic test set.
In particular, we compare it with a baseline U-Net (with 5 downsampling/upsampling blocks) that takes a stack of all the coarsely aligned images ($H_{i,j} \forall j$  in Eqn.~\ref{eqn:encoder-input}) as input for its encoder, and skip connections from the encoder to the four SVBRDF decoders.
This architecture has been widely used for SVBRDF estimation~\cite{deschaintre2018single, li2018materials, li2018learning}.
As can be seen in Tab.~\ref{table:comp-synthetic}, while our diffuse albedo prediction is slightly ($1.7\%$) worse than the U-Net we significantly outperform it in specular albedo, roughness and normal predictions, 
with $31\%$, $23\%$ and $9.5\%$ lower $L_2$ loss respectively.
This is in spite of not using skip-connections in our network (to allow for optimization later in our pipeline).
We also compare our results with the state-of-the-art single-image shape and SVBRDF estimation method of Li et al.~\cite{li2018learning}.
Unsurprisingly, we outperform them significantly, demonstrating the usefulness of aggregating multi-view information.

\begin{table}[t]
\renewcommand{\arraystretch}{1.2}
\setlength{\tabcolsep}{3.0pt}
\begin{tabular}{lcccc}
    \hline

    &  {Diffuse} & {Normal} &  {Roughness}  & {Specular} \\ \hline
    Naive U-Net &   \textbf{0.0060}     &  0.0336      &  0.0359         &  0.0125 \\
    \textbf{Ours}   & 0.0061     &  \textbf{0.0304} &    \textbf{0.0275}       &   \textbf{0.0086}              \\ \hhline{=====}
    Li et al.~\cite{li2018learning}   & 0.0227 &  0.1075      &  0.0661        & ---                \\
    \textbf{Ours (\small{$256 \times 256$})} &   \textbf{0.0047}    &  \textbf{0.0226}      &  \textbf{0.0257}         &     \textbf{0.0083}          
\\ \hline
\end{tabular}
\caption{Quantitative SVBRDF evaluation on a synthetic test set. 
    We report the $L_2$ error. Since Li et al.~\cite{li2018learning} work on 
    $256 \times 256$ images, we downsample and evaluate at that resolution.
    Also, they do not predict the specular albedo.}
\label{table:comp-synthetic}
\vspace{\spaceunderfigures}
\end{table}


\subsection{Evaluation on Real Captured Data}

We evaluate our method on real data captured using a gantry with a FLIR camera and a nearly collocated light to mimic our capture setup. \textbf{Please refer to the supplementary material for additional results.}

\begin{figure}[t]
    \centering
    \includegraphics[width=\linewidth]{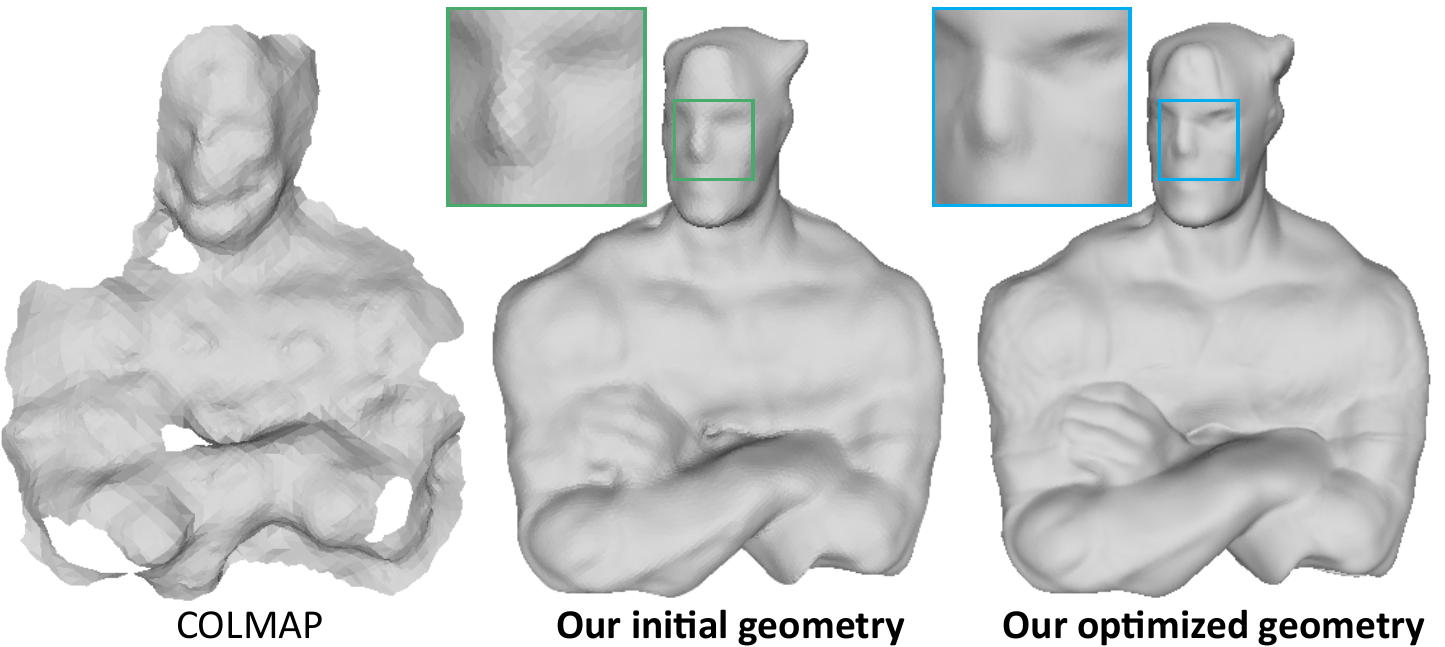}
    \caption{Comparison on geometry reconstruction. COLMAP fails to reconstruct a complete mesh from the sparse 
    inputs. In contrast, our initial mesh is of much higher quality, and our joint optimization recovers even more fine-grained details on the mesh. Input image in Fig.~\ref{fig:real-results} (top).}
    \label{fig:comp-colmap}
\end{figure}

\boldstartspace{Evaluation of geometry reconstruction.}
Our framework combines our predicted depths and normals to reconstruct the initial mesh.
Figure~\ref{fig:comp-colmap} shows the comparison between our reconstructed mesh and
the mesh from COLMAP, a state-of-the-art multi-view stereo framework \cite{schoenberger2016mvs}.
From such sparse inputs and low-texture surfaces, COLMAP is not able to find reliable correspondence across views,
which results in a noisy, incomplete 3D mesh. 
In contrast, our initial mesh is already more complete and detailed, as a result of our more accurate depths and normals. 
Our joint optimization further refines the per-vertex normals and extracts fine-scale detail in the object geometry. 

\begin{figure}[t]
    \centering
    \includegraphics[width=\linewidth]{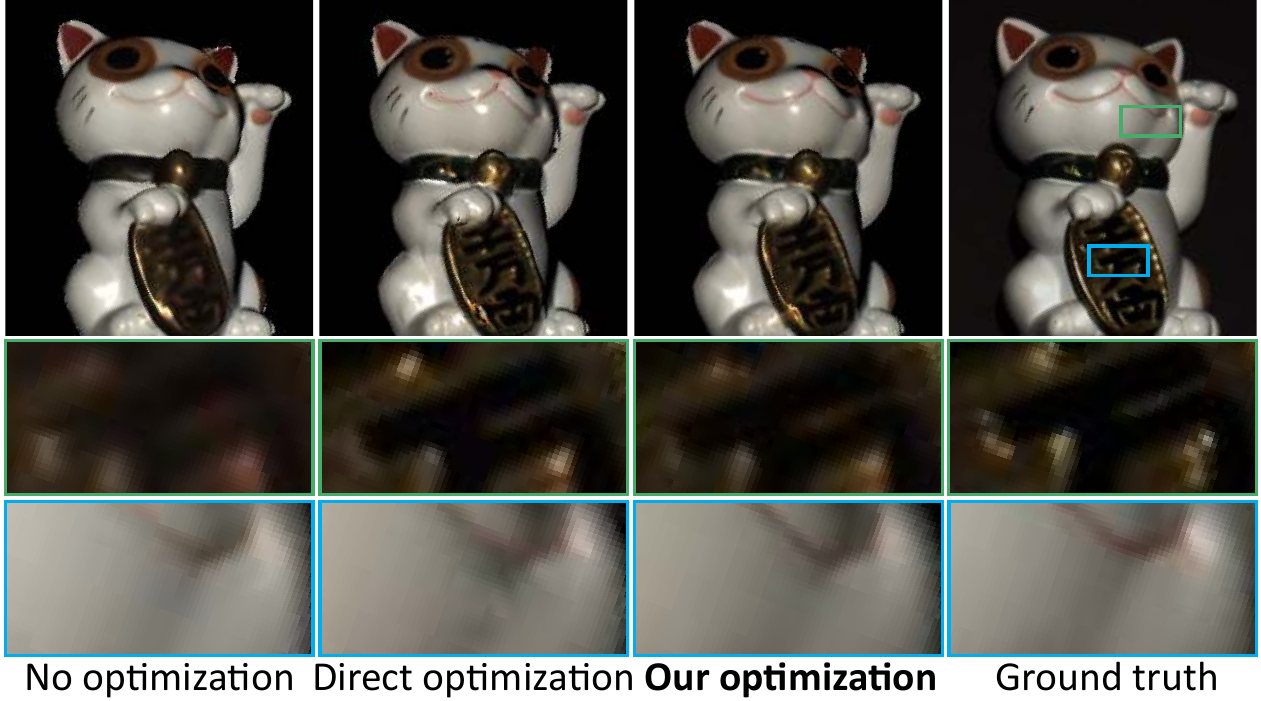}
    \caption{Comparison on SVBRDF optimization. Simple averaging without optimization produces blurry 
    results, and direct per-vertex optimization results in outliers and discontinuities. In comparison, 
    our optimization generates more visually plausible results.}
    \label{fig:comp-direct-opt}
    \vspace{\spaceunderfigures}
\end{figure}

\boldstartspace{Evaluation of SVBRDF optimization.}
We compare our SVBRDF and geometry optimization scheme (Sec.~\ref{sec:opt}) 
with averaging the per-view predictions using weights based on the angle between the viewpoint and surface normal, as well as this averaging followed by per-vertex optimization. 
From Fig.~\ref{fig:comp-direct-opt} we can see that the weighted averaging produces blurry results. Optimizing the per-vertex BRDFs  brings back detail but also has spurious discontinuities in appearance because of the lack of any regularization.
In contrast, our latent-space optimization method recovers detailed appearance without these artifacts.

\begin{figure}[t]
    \centering
    \includegraphics[width=\linewidth]{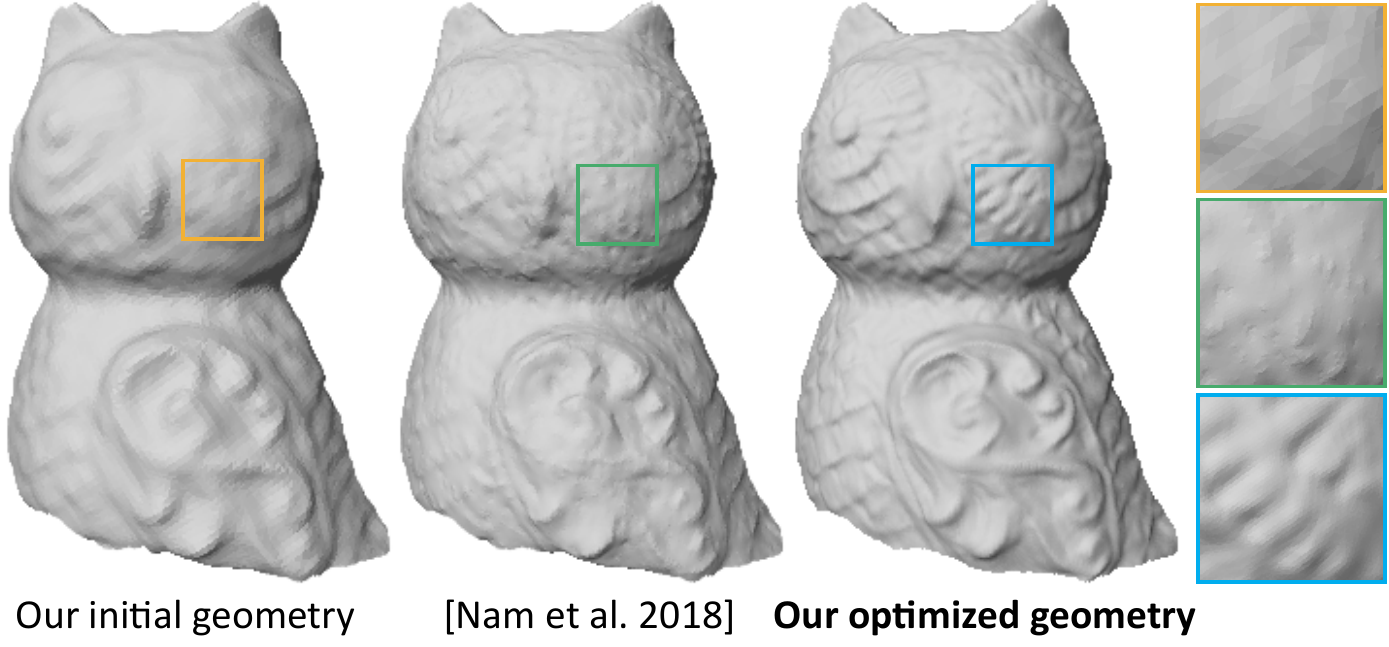}
    \caption{Comparison with Nam et al.~\cite{nam2018practical}. While both have the same initialization, our learning-based refinement produces more accurate, detailed geometry. Input in Fig.~\ref{fig:real-results}.}
    \label{fig:comp-nam-geo}
    \vspace{\spaceunderfigures}
\end{figure}

\begin{figure*}[t]
    \centering
    \includegraphics[width=\linewidth]{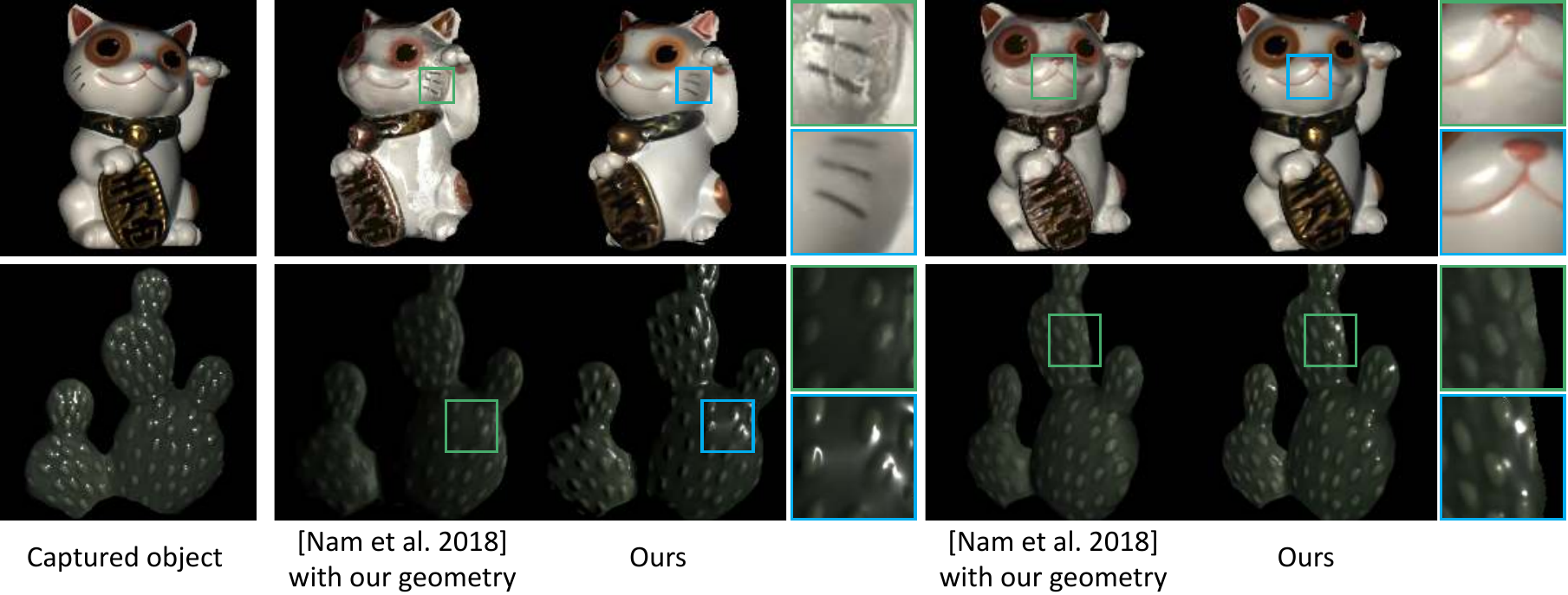}
    \caption{Comparison with Nam et al.~\cite{nam2018practical}. We render two reconstructed objects under \emph{novel} viewpoints and lighting. Nam et al. are not able to accurately reconstruct appearance from sparse views, and produce noisy edges and incorrect specular highlights (top) or miss the specular component completely (bottom). In contrast, our method produces photorealistic results. }
    \label{fig:comp-nam}
    \vspace{-0.4cm}
\end{figure*}

\begin{figure*}[t]
    \centering
    \includegraphics[width=\linewidth]{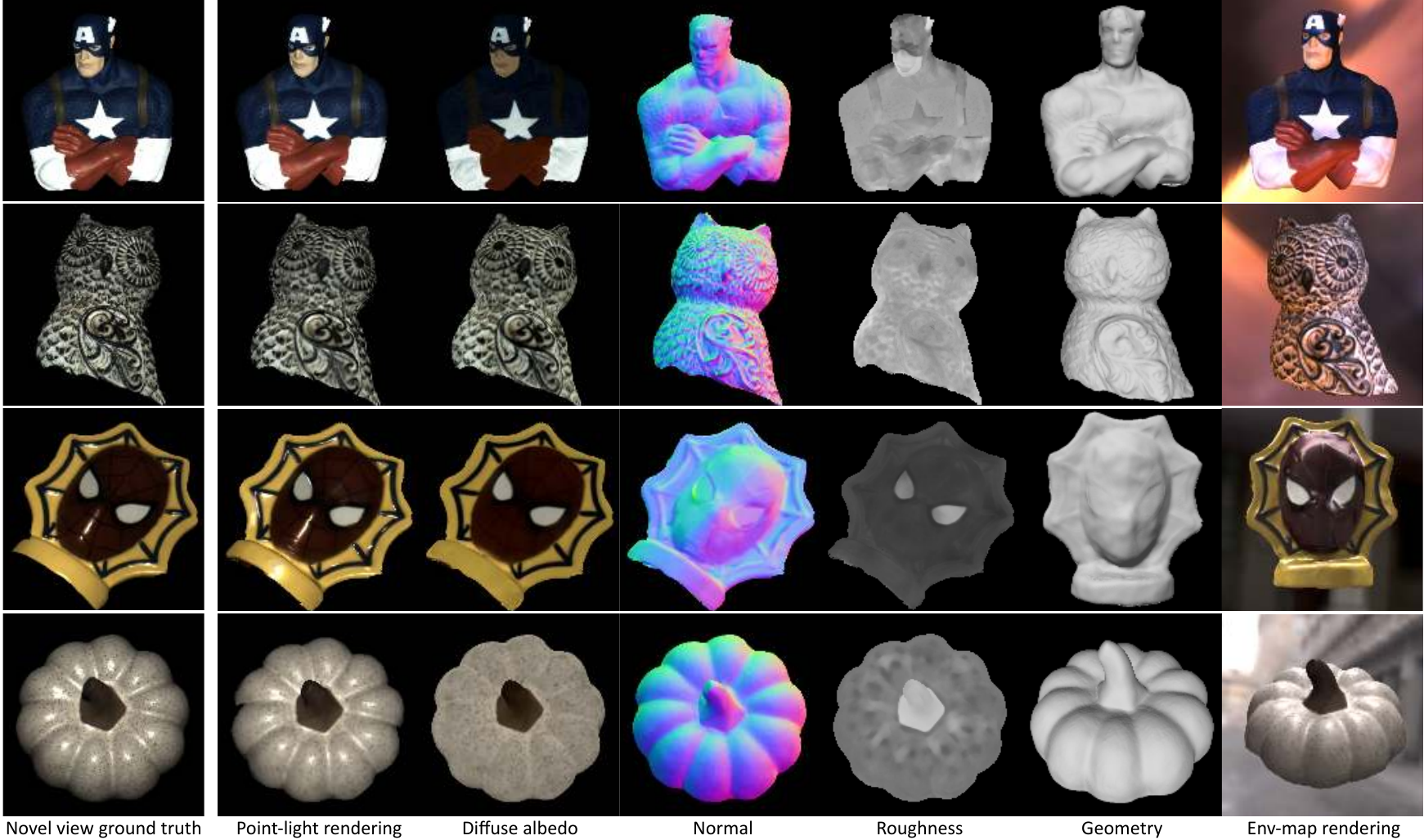}
    \caption{Results on real scenes. For each scene, we show our reconstructed geometry, normal map and 
        SVBRDF components (please refer to supplementary materials for specular albedo).  
        We compare our point-light rendering results (second column) under \emph{novel} viewpoints and lighting with captured
        ground truth photographs (first column). We also show a rendering of the object with our 
        reconstructed appearance under environment lighting (last column).}
    \label{fig:real-results}
    \vspace{-0.4cm}
\end{figure*}

\boldstartspace{Comparisons against Nam et al.~\cite{nam2018practical}}
We also compare our work with the state-of-the-art geometry and reflectance reconstruction method of Nam et al.
Their work captures 60+ images of an object with a handheld camera under collocated lighting;  
they first use COLMAP~\cite{schoenberger2016mvs} to reconstruct the coarse shape and use it to bootstrap a physics-based optimization process to recover per-vertex normals and BRDFs.
COLMAP cannot generate complete meshes from our sparse inputs (see Fig.~\ref{fig:comp-colmap}).
Therefore, we provided our input images, camera calibration, \emph{and initial geometry} to the authors who processed this data. 
As can be seen in Fig.~\ref{fig:comp-nam-geo}, our final reconstructed geometry has significantly more details than their final optimized result in spite of starting from the same initialization.
Since they use a different BRDF representation than ours, making direct SVBRDF comparisons difficult, in Fig.~\ref{fig:comp-nam} we compare renderings of the reconstructed object under novel lighting and viewpoint.
These results show that they cannot handle our sparse input and produce noise, erroneous reflectance (\textsc{Cat} scene) or are unable to recover the specular highlights of highly specular objects (\textsc{Cactus}) scene. 
In comparison, our results have significantly higher visual fidelity. Please refer to the supplementary video for more renderings. 

\boldstartspace{More results on real data.}
Figure~\ref{fig:real-results} shows results from our method on additional real scenes. 
We can see here that our method can reconstruct detailed geometry and appearance for objects with a wide variety of complex shapes and reflectance. Comparing renderings of our estimates under novel camera and collocated lighting against ground truth captured photographs demonstrates the accuracy of our reconstructions. 
We can also photorealistically render these objects under novel environment illumnination.
Please refer to the supplementary document and video for more results.

\boldstartspace{Limitations.} Our method might fail to handle highly non-convex objects, where some parts are
visible in as few as a single view and there are no correspondence cues to infer correct depth.
In addition, we do not consider global illumination in SVBRDF optimization. While it is a reasonable approximation
in most cases, it might fail in some particular scenes with strong inter-reflections. 
For future work, it would be interesting to combine our method with physics-based differentiable rendering~\cite{li2018diff, zhang2019diff} 
to handle these complex light transport effects.

%% file: Conclusions.tex
\vspace{-0.2cm}
\section{Conclusion}\label{sec:conclusions}
We have proposed a learning-based framework to reconstruct the geometry and appearance of an arbitrary 
object
from a sparse set of just six images. 
We predict per-view depth using learning-based MVS, and design a novel multi-view reflectance estimation network 
that robustly aggregates information from our sparse views for accurate normal and SVBRDF estimation.
We further propose a novel joint optimization in latent feature space to fuse and refine our multi-view 
predictions. 
Unlike previous methods that require densely sampled images, our method produces high-quality reconstructions from a sparse set of images, and presents a step towards practical appearance capture for 3D scanning and VR/AR applications.  

\boldstart{Acknowledgements}
This work was supported in part by NSF grant 1617234, ONR grants
N000141712687, N000141912293, Adobe and the UC San Diego Center for
Visual Computing.

%% file: network.tex
\section{BRDF Model}
We use a simplified version of the Disney BRDF model~\cite{burley2012disney} proposed by Karis et al.~\cite{karis2013real}. 
Let $A$, $N$, $R$, $S$ be the diffuse albedo, normal, roughness and specular albedo respectively,  
$L$ and $V$ be the light and view direction, and $H = \frac{V + L}{2}$ be their half vector. 
Our BRDF model is defined as: 
\begin{equation}
f(A, N, R, L, V) = \frac{A}{\pi} + \frac{D(H, R)F(V, H, S)G(L, V, H, R)}{4(N\cdot L)(N\cdot V)}
\end{equation}
where $D(H, R)$, $F(V, H, S)$ and $G(L, V, H, R)$ are the \emph{normal distribution}, \emph{fresnel} and 
\emph{geometric terms} respectively. These terms are defined as follows:
\begin{eqnarray}
D(H, R) &=& \frac{\alpha^{2}}{\pi\left[(N\cdot H)^{2}(\alpha^{2} - 1) + 1\right]^{2} } \nonumber  \\
\alpha &=& R^{2} \nonumber \\
F(V, H, S) &=& S + (1 - S ) 2^{ -\left[5.55473(V\cdot H) + 6.8316\right](V\cdot H)} \nonumber \\
G(L, V, R) &=& G_{1}(V, N)G_{1}(L, N) \nonumber \\ 
G_{1}(V, N) &=& \frac{N\cdot V}{(N\cdot V)(1 -k) + k} \nonumber \\
G_{1}(L, N) &=& \frac{N\cdot L}{(N\cdot L)(1 -k) + k} \nonumber \\
k &=& \frac{(R + 1)^2}{8} \nonumber 
\end{eqnarray}

\begin{figure*}[t]
    \centering
    \includegraphics[width=\textwidth]{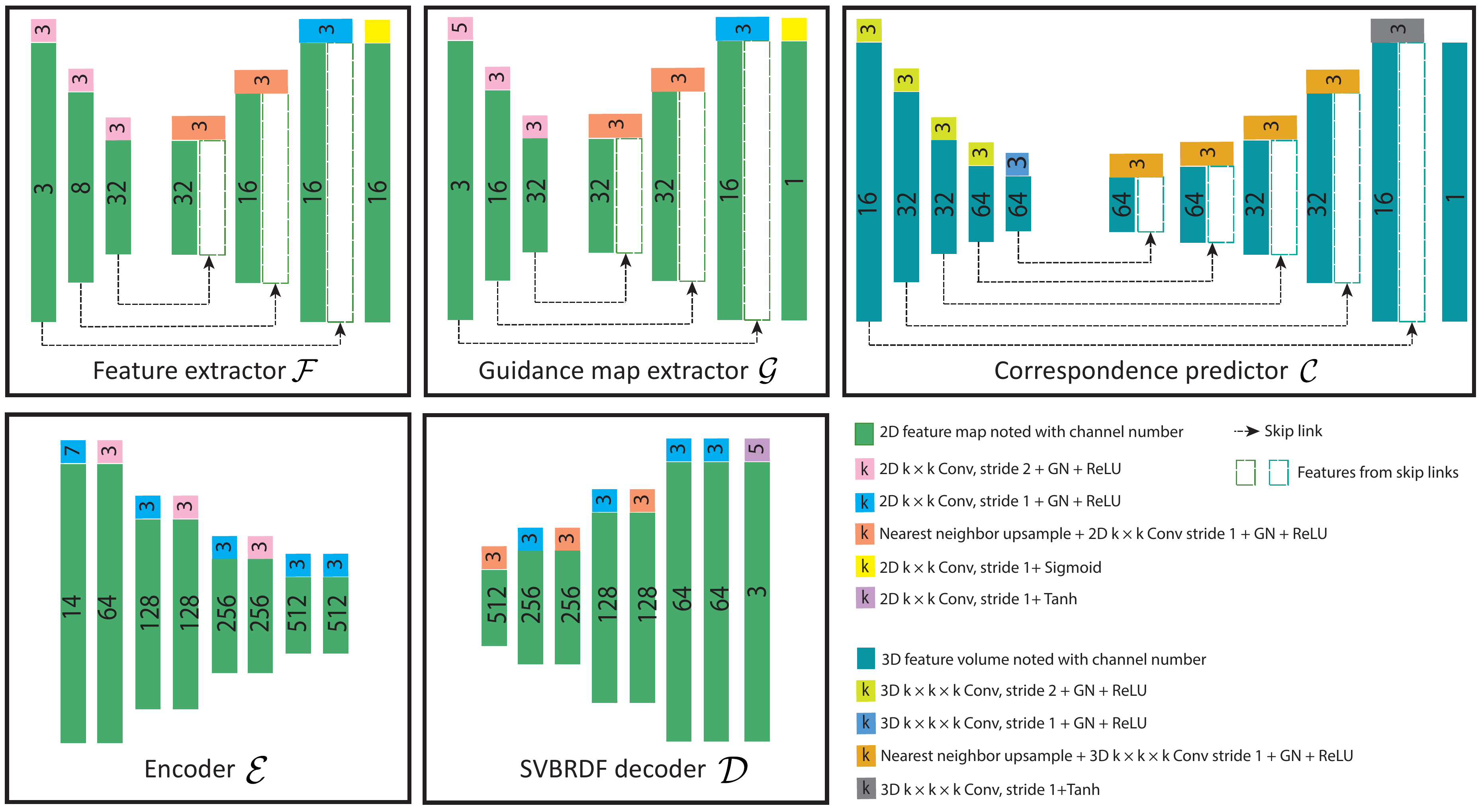}
    \caption{Our network architecture.}
    \label{fig:network}
\end{figure*}
\section{Network Architecture}\label{sec:network}
We have talked about the motivations, design and core components of 
our depth prediction network and SVBRDF prediction network in Sec.~3.1 and Sec.~3.2 in the paper.
We now introduce the network architectures in detail as shown in Fig.~\ref{fig:network}.

\boldstartspace{Depth prediction network.}
As discussed in Sec.~3.1 in the paper, the depth prediction network consists of three parts: 
the feature extractor $\mc{F}$, the correspondence predictor $\mc{C}$ and the guidance map extractor $\mc{G}$.
The feature extractor $\mc{F}$ and the correspondence predictor $\mc{C}$ are used to predict the initial depth map $D_i'$; 
the guidance map extractor is applied to refine $D_i'$ using a guided filter~\cite{wu2018fast} to obtain the final depth $D_i$.
Figure~\ref{fig:network} shows the details of these sub-networks in the first row.

We use the feature extractor and the correspondence predictor to regress the initial depth, similar to \cite{xu2019deep}. 
In particular, the feature extractor $\mc{F}$ is a 2D U-Net that consists of 
multiple downsampling and upsampling convolutional layers with skip links, group normalization (GN)~\cite{wu2018group} 
layers and ReLU activation layers; it extracts per-view image feature maps with 16 channels.

To predict the depth $D_i$ at reference view $i$, we uniformly sample 128 frontal parallel depth planes at depth $d_1, d_2, \ldots, d_{128}$ 
in front of that view within a pre-defined depth range $[d_1, d_{128}]$ that covers the target object we want to capture.
We project the feature maps from all views onto every depth plane at view $i$ using homography-based warping 
to construct the plane sweep volume of view $i$. We then build a cost volume by calculating 
the variance of the warped feature maps over views at each plane.
The correspondence predictor $\mc{C}$ is a 3D U-Net that processes this cost volume; 
it has multiple downsampling and upsampling 3D convolutional layers with skip links, GN layers and ReLU layers.
The output of $\mc{C}$ is a 1-channel volume, and we apply soft-max on this volume across the depth planes 
to obtain the per-plane depth probability maps $P_1, \ldots, P_{128}$ of the $128$ depth planes; 
these maps indicate the probability of the depth of a pixel being the depth of each plane.
A depth map is then regressed by linearly combining the per-plane depth values weighted by the per-plane depth probability maps:
\begin{align}
    D_i'=\sum_{q=1}^{128} P_q*d_q.
    \label{eqn:depth}
\end{align}
We apply the guidance map extractor $\mc{G}$ to refine the initial depth $D_i'$.
$\mc{G}$ is a 2D U-Net that outputs a 1-channel feature map.
We use the output feature map as a guidance map to filter the initial depth $D_i'$ and 
obtain the final depth $D_i$.

\boldstartspace{SVBRDF prediction network.}
We have discussed the SVBRDF prediction network in Sec.~3.2, and shown the overall architecture, input and output in Fig.~2 and Fig.~3 of the paper.
We now introduce the details of the encoder $\mc{E}$ and the SVBRDF decoder $\mc{D}$ in Fig.~\ref{fig:network} (bottom row).
Specifically, the encoder consists of a set of convolutional layers, followed by GN and ReLU layers; 
multiple convolutional layers with a stride of $2$ are used to downsample the feature maps three times.
The decoder upsamples the feature maps three times with nearest-neighbor upsampling, 
and applies convolutional layers, GN and ReLU layers to process the feature maps at each upsampling level.
As discussed in Sec.~3.2 of the paper, 
we apply four decoders with the same architecture, which are connected with the same encoder, 
to regress three BRDF components and the normal map at each input view.

\section{Comparison on SVBRDF Prediction}\label{svbrdf}
In Sec.~4.1 and Tab.~1 of the paper, we have shown quantitative comparisons on synthetic data 
between our network, the na\"ive U-Net and a single-image SVBRDF prediction network proposed by Li et al.~\cite{li2018learning}.
We now demonstrate qualitative comparisons between these methods on both synthetic 
and real examples in Fig.~\ref{fig:compsyn1}, Fig.~\ref{fig:compsyn2}, Fig.~\ref{fig:compreal1} and Fig.~\ref{fig:compreal2}.
From these figures, we can see that the na\"ive U-Net produces noisy normals and the single-view method \cite{li2018learning} 
produces normals with very few details, whereas our predicted normals are of much higher quality, especially
in regions where there are serious occlusions (indicated by the \emph{red} arrow).
In contrast, as reflected by the comparison on synthetic data in Fig.~\ref{fig:compsyn1} and Fig.~\ref{fig:compsyn2}, 
our predictions are more accurate and more consistent with the ground truth than the other methods.
These results demonstrate that our novel network architecture (see Sec.~3.2 in the paper) allows for effective
aggregation of multi-view information and leads to high-quality per-view SVBRDF estimation.


\section{Comparison on Geometry Reconstruction}\label{svbrdf}
\begin{figure*}[t!]
    \centering
    \includegraphics[width=\textwidth]{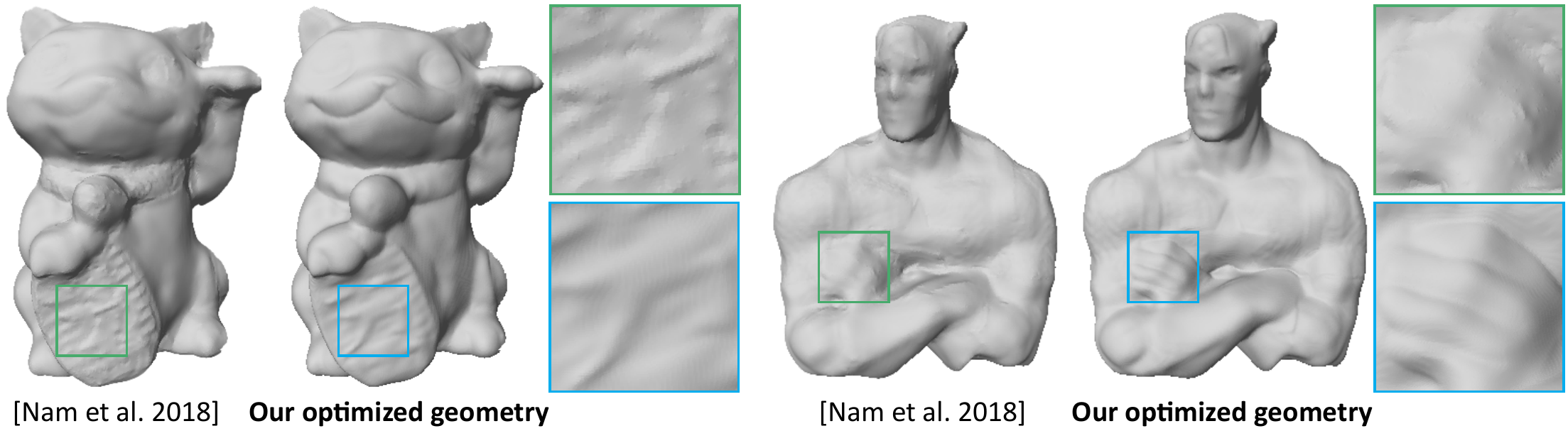}
    \caption{Comparison with Nam et al.~\cite{nam2018practical} on geometry optimization. Our results have more fine-grained details and fewer artifacts.}
    \label{fig:geo-opt-comp}
\end{figure*}

In Fig.~6 of the paper, we compare our optimized geometry against the optimized result from Nam et al.~\cite{nam2018practical} 
that uses the same initial geometry as ours.
We show additional comparisons on real data in Fig.~\ref{fig:geo-opt-comp}.
Similar to the comparison in the paper, our optimized geometry is of much higher quality 
than Nam et al. with more fine-grained details and fewer artifacts.

\section{Additional Ablation Study}\label{svbrdf}
In this section, we demonstrate additional experiments to justify the design choices in our pipeline,
including input variants of the SVBRDF estimation network, non-rigid warping and per-vertex refinement.

\begin{figure}[hhh!]
    \centering
    \includegraphics[width=\linewidth]{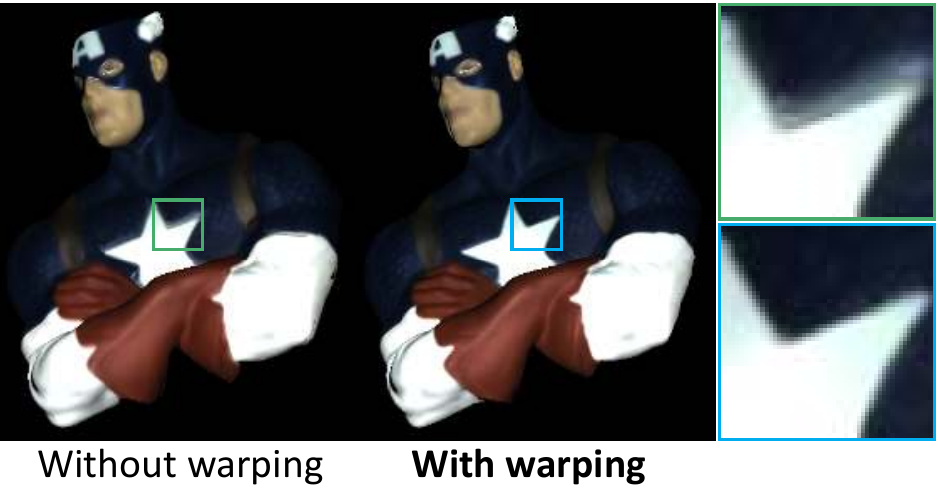}
    \caption{Comparison between optimizations with and without per-view warping. Our method with warping removes the ghosting artifacts around the edges.}
    \label{fig:warp-comp}
\end{figure}

\begin{table}[t]
\renewcommand{\arraystretch}{1.2}
\setlength{\tabcolsep}{1.0pt}
\begin{tabular}{lcccc}
    \hline
    {Network input} &  {Diffuse} & {Normal} &  {Roughness}  & {Specular} \\ \hline
    $I_{i \la j}$ &   0.0081     &  0.0456     &  0.0379        &  0.0098 \\
    $I_i, I_{i \la j}$   & 0.0071     &  0.0363 &    0.0304       &   0.0109        \\   
    $I_i, I_{i \la j}, Z_{i \la j}, Z^*_{i \la j}$  & 0.0063 &  0.0321      &  0.0306       &  0.0098                \\
    $I_i, I_{i \la j}, L_{i}, L_{i \la j}$  & \textbf{0.0061} &  \textbf{0.0304}      &  0.0299        & 0.0093               \\
    \textbf{Ours full}   & \textbf{0.0061}     &  \textbf{0.0304} &    \textbf{0.0275}       &   \textbf{0.0086}              \\ \hline
\end{tabular}
\caption{Quantitative comparisons between networks trained with different inputs on the synthetic test set.  }
\label{table:comp-synthetic}
\end{table}

\begin{figure}[t!]
    \centering
    \includegraphics[width=\linewidth]{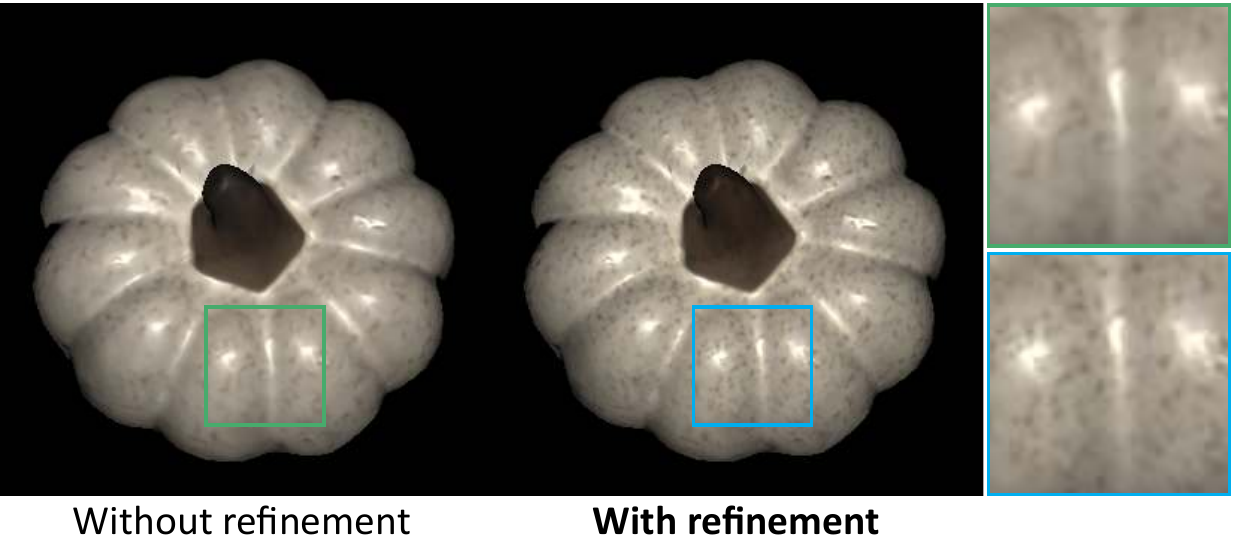}
    \caption{Comparison on results with and without per-vertex refinement. With the refinement, our method is able to recover high-frequency details such as the spots on the object.}
    \label{fig:pervertex-opt-comp}
\end{figure}
\boldstartspace{Network inputs.}
Our SVBRDF network considers the input image ($I_i$), the warped images ($I_{i \la j}$), the light/viewing (which are collocated) direction maps ($L_i$ and $L_{i\la j}$), and the depth maps ($Z_{i \la j}$ and $Z^*_{i \la j}$) as inputs (please refer to Sec.~3.2 in the paper for details of these input components).
We verify the effectiveness of using these inputs by training and comparing multiple networks with different subsets of the inputs.
In particular, we compare our full model against a network that uses only the warped image $I_{i \la j}$, a network that considers both $I_{i \la j}$ and the reference image $I_i$, a network that uses the reference image, warped image and the depth, and a network that uses 
the reference image, warped image, and the viewing directions.
Table.~\ref{table:comp-synthetic} shows the quantitative comparisons between these networks on the synthetic testing set. 
The network using a pair of images ($I_i$, $I_{i \la j}$) improves the accuracy for most of the  terms over the one that uses only the warped image ($I_{i \la j}$), which reflects the benefit of involving multi-view cues in the encoder network.
On top of the image inputs, the two networks that involve additional depth information ($Z_{i \la j}$, $Z^*_{i \la j}$) and the viewing directions ($L_i$, $L_{i\la j}$) both obtain better performance than the image-only versions, which leverage visibility cues and photometric cues from the inputs respectively.
Our full model is able to leverage both cues from multi-view inputs and 
achieves the best performance.

\boldstartspace{Per-view warping.} Due to potential inaccuracies in the geometry, 
the pixel colors of a vertex from different views may not be consistent. Directly minimizing 
the difference between the rendered color and the pixel color of each view will lead to ghosting artifacts, as shown 
in Fig.~\ref{fig:warp-comp}. To solve this problem, we propose to apply a non-rigid warping to each view. 
From Fig.~\ref{fig:warp-comp} we can see that non-rigid warping can effectively tackle the misalignments and 
leads to sharper edges.

\boldstartspace{Per-vertex refinement.}  As shown in Fig.~\ref{fig:pervertex-opt-comp}, the 
image rendered using estimated SVBRDF without per-vertex refinement loses high-frequency
details such as the tiny spots on the pumpkin, due to the existence of the bottleneck in 
our SVBRDF network. In contrast, the proposed per-vertex refinement can successfully recover these details and reproduces more faithful appearance of the object.

\begin{figure*}[t]
    \centering
    \includegraphics[width=\textwidth]{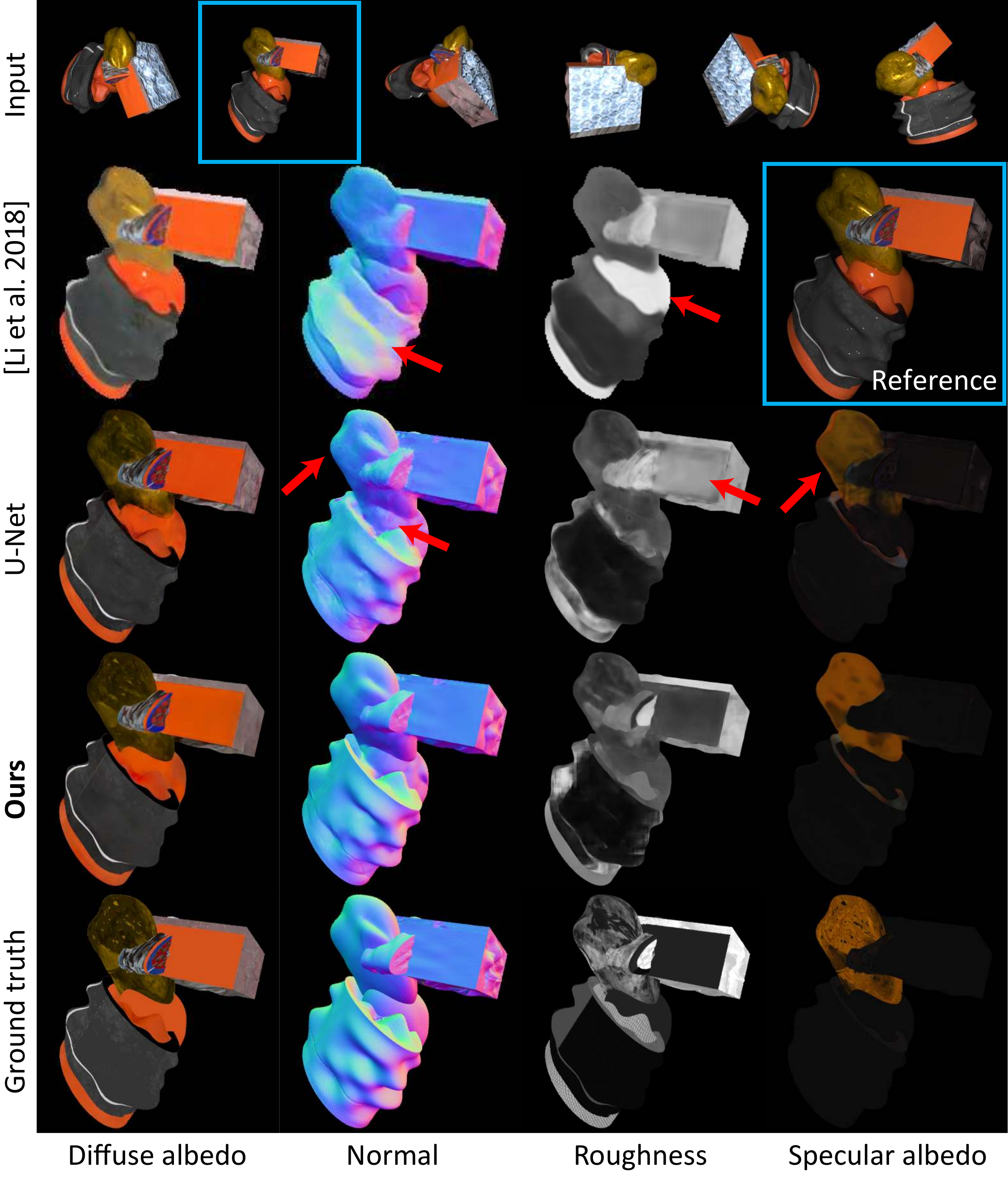}
    \caption{Qualitative comparison of SVBRDF estimation on synthetic data. Note that Li et al.~\cite{li2018learning} do not predict specular albedo.}
    \label{fig:compsyn1}
\end{figure*}

\begin{figure*}[t]
    \centering
    \includegraphics[width=\textwidth]{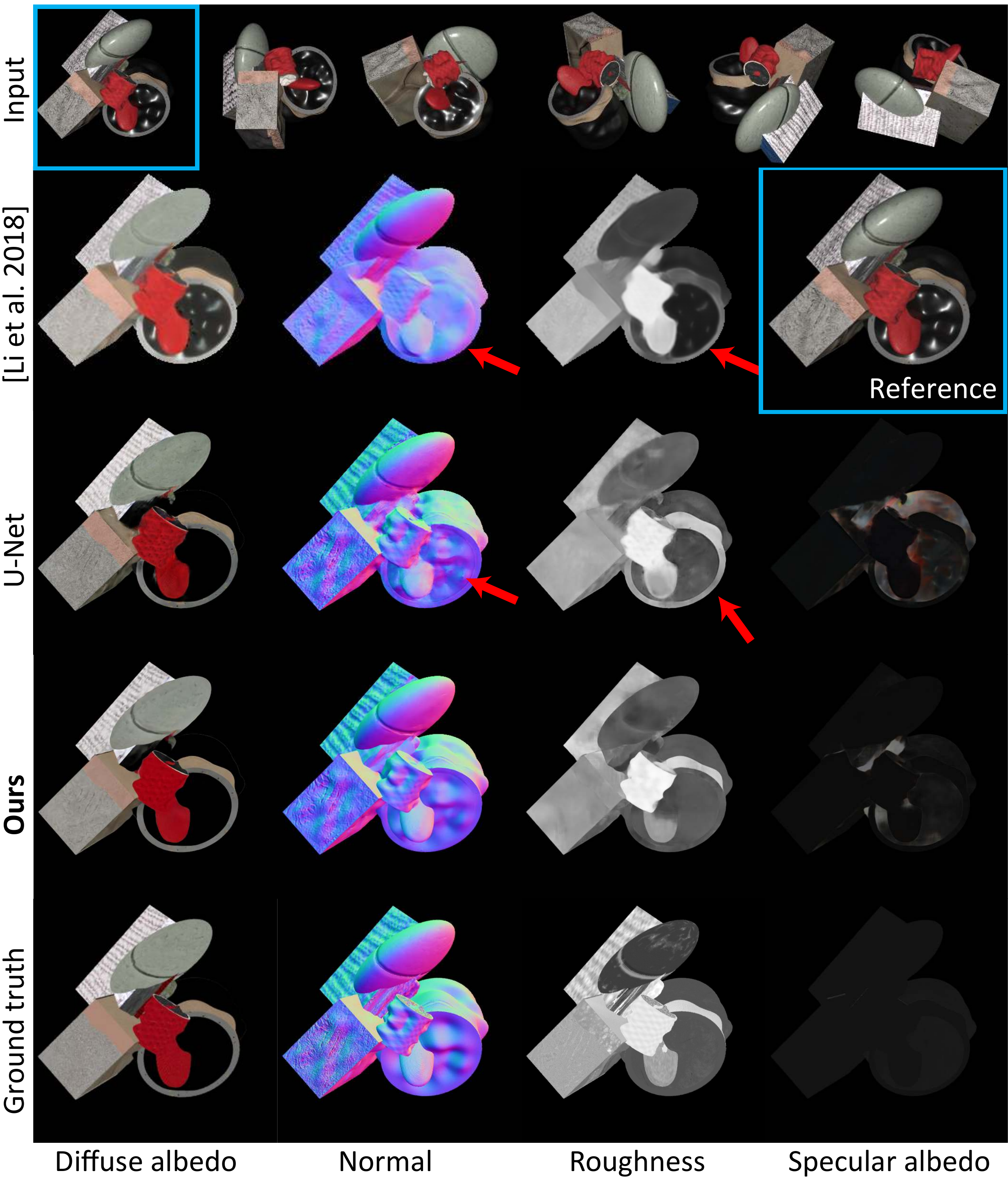}
    \caption{Qualitative comparison of SVBRDF estimation on synthetic data. Note that Li et al.~\cite{li2018learning} do not predict specular albedo.}
    \label{fig:compsyn2}
\end{figure*}

\begin{figure*}[t]
    \centering
    \includegraphics[width=\textwidth]{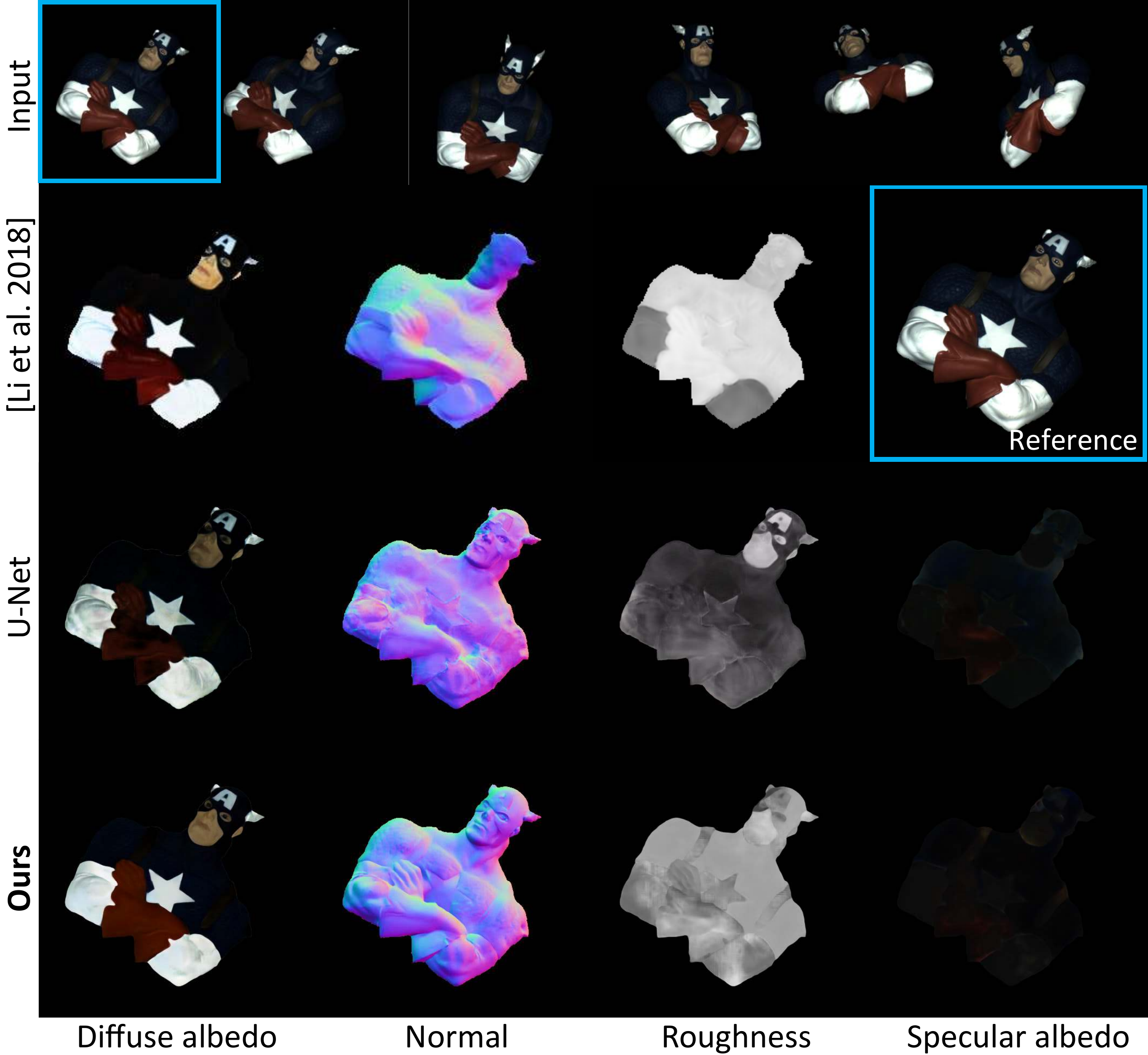}
    \caption{Qualitative comparison of SVBRDF estimation on real data. Note that Li et al.~\cite{li2018learning} do not predict specular albedo.}
    \label{fig:compreal1}
\end{figure*}

\begin{figure*}[t]
    \centering
    \includegraphics[width=\textwidth]{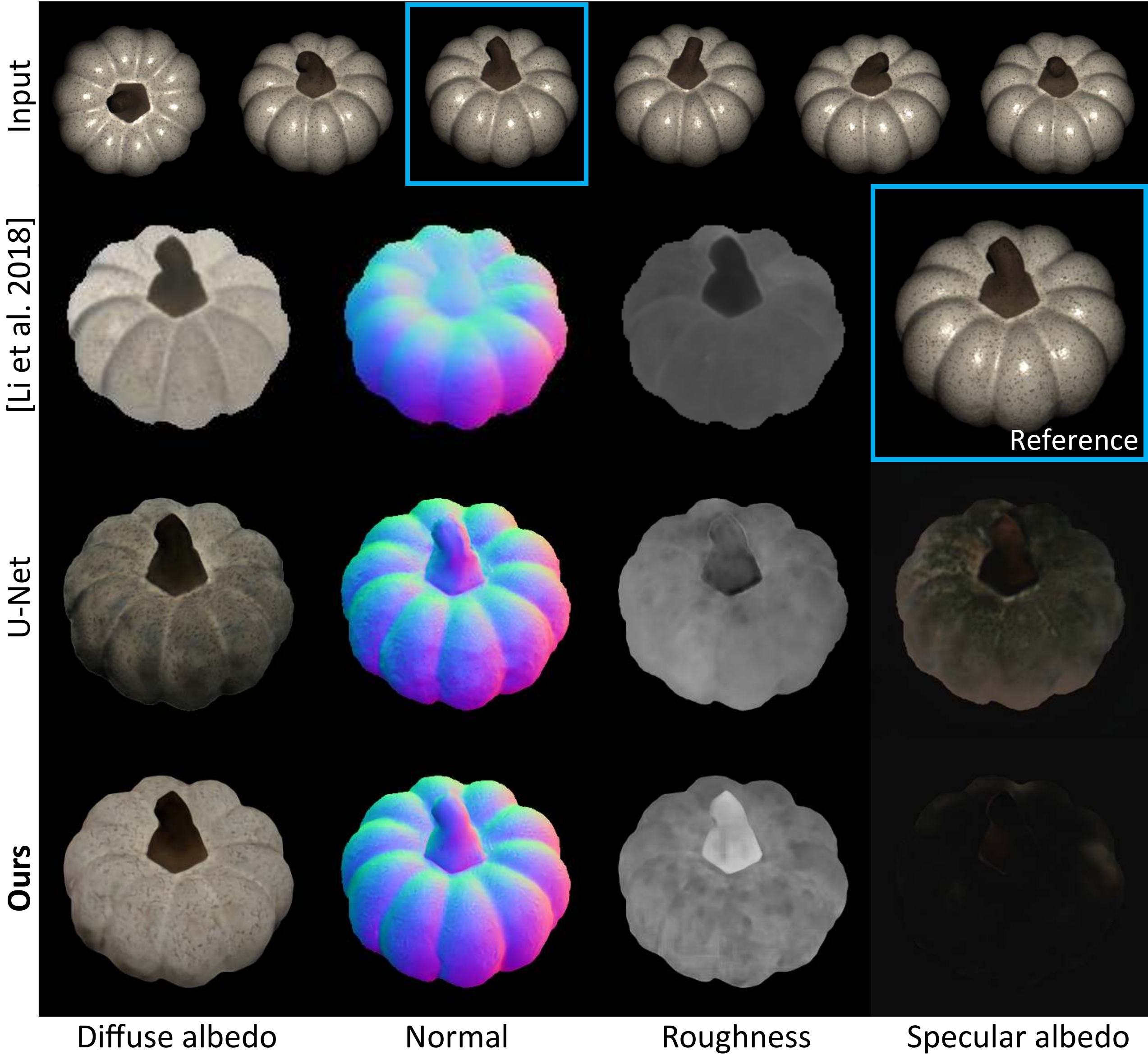}
    \caption{Qualitative comparison of SVBRDF estimation on real data. Note that Li et al.~\cite{li2018learning} do not predict specular albedo.}
    \label{fig:compreal2}
\end{figure*}